\documentclass[letterpaper]{article} 
\usepackage{aaai25}  
\usepackage{times}  
\usepackage{helvet}  
\usepackage{courier}  
\usepackage[hyphens]{url}  
\usepackage{graphicx} 
\usepackage{natbib}  
\usepackage{caption} 
\usepackage{algorithm}
\usepackage{algorithmic}
\usepackage{amsmath} 
\usepackage{bbding}
\usepackage{makecell}
\usepackage{xcolor}
\usepackage{xpatch}
\usepackage{colortbl}  
\usepackage{array} 
\usepackage{booktabs, multirow, tabularx}

\usepackage{newfloat}
\usepackage{listings}
\DeclareCaptionStyle{ruled}{labelfont=normalfont,labelsep=colon,strut=off} 
\floatstyle{ruled}
\newfloat{listing}{tb}{lst}{}
\floatname{listing}{Listing}
\title{Guided Real Image Dehazing using YCbCr Color Space}
\author{
    Wenxuan Fang\textsuperscript{\rm 1}, Junkai Fan\textsuperscript{\rm 1}, Yu Zheng\textsuperscript{\rm 1}, Jiangwei Weng\textsuperscript{\rm 1}, Ying Tai\textsuperscript{\rm 2}, Jun Li\textsuperscript{\rm 1}\thanks{Corresponding authors}
}
\affiliations{
    \textsuperscript{\rm 1}School of Computer Science and Engineering, Nanjing University of Science and Technology, Nanjing, China\\
    \textsuperscript{\rm 2}School of Intelligence Science and Technology, Nanjing University, Suzhou, China.\\

    \{wenxuan\_fang, junkai.fan, yuzheng, wengjiangwei, junli\}@njust.edu.cn, yingtai@nju.edu.cn
}

\usepackage{bibentry}

\begin{document}

\maketitle

\begin{abstract}
Image dehazing, particularly with learning-based methods, has gained significant attention due to its importance in real-world applications. However, relying solely on the RGB color space often fall short, frequently leaving residual haze. This arises from two main issues: the difficulty in obtaining clear textural features from hazy RGB images and the complexity of acquiring real haze/clean image pairs outside controlled environments like smoke-filled scenes. To address these issues, we first propose a novel \textbf{S}tructure \textbf{G}uided \textbf{D}ehazing \textbf{N}etwork (SGDN) that leverages the superior structural properties of YCbCr features over RGB. It comprises two key modules: Bi-Color Guidance Bridge (BGB) and Color Enhancement Module (CEM). BGB integrates a phase integration module and an interactive attention module, utilizing the rich texture features of the YCbCr space to guide the RGB space, thereby recovering clearer features in both frequency and spatial domains. To maintain tonal consistency, CEM further enhances the color perception of RGB features by aggregating YCbCr channel information. Furthermore, for effective supervised learning, we introduce a Real-World Well-Aligned Haze (RW$^2$AH) dataset, which includes a diverse range of scenes from various geographical regions and climate conditions. Experimental results demonstrate that our method surpasses existing state-of-the-art methods across multiple real-world smoke/haze datasets. Code and Dataset: \textcolor{blue}{\url{https://github.com/fiwy0527/AAAI25_SGDN.}}
\end{abstract}

\section{Introduction}
\label{intro}
Image dehazing aims to remove the blurring effect caused by atmospheric particles such as haze and smoke in images and improve image quality, which has important application value in object detection \cite{zhao2024detrs}, image segmentation \cite{segment} and object tracking \cite{hu2023transformer, hu2024towards}. The formation of foggy images can usually be modeled using an Atmospheric Scattering Model (ASM):
\begin{equation}
    I(x) = J(x)t(x)+A(x)(1-t(x)),
\label{eq:eq1}
\end{equation}
where $I(x)$ and $J(x)$ represent the observed foggy image and clean image, and $A(x)$ and $t(x)$ are the global atmospheric light and transmission map, respectively. The transmission map $t(x)=e^{-\beta d(x)}$ is determined by the scattering coefficient $\beta$ and the scene depth $d(x)$. 
Many dehazing methods \cite{aodnet, yoly} use deep networks to estimate physical parameters and apply Eq.~\eqref{eq:eq1} to restore clean images. Additionally, some methods \cite{sdbad, DEANet} focus on learning a mapping model from hazy to clean images in a supervised manner, offering an end-to-end dehazing solution. Despite their promising results on synthetic haze datasets, they face two major challenges in real-world haze scenes.
\begin{figure}[!t]
    \centering 
    \includegraphics[width=0.97\linewidth]{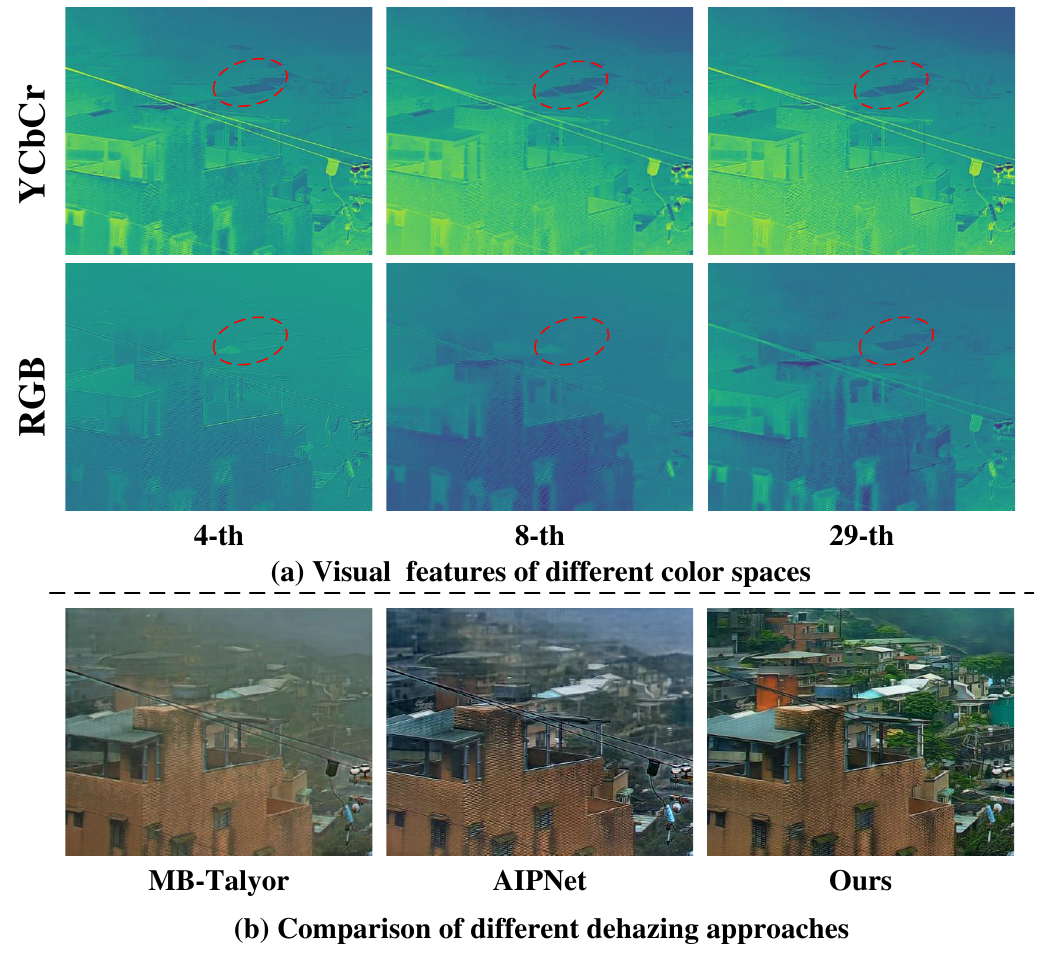}
    \caption{Visual comparison of different color spaces: (a) RGB features degrade, blurring textures, while YCbCr is less affected by fog and shows clearer textures. (b) RGB models (e.g., MB-Taylor) leave residual haze, YCbCr models (e.g., AIPNet) distort colors. Our approach removes heavy fog while preserving color accuracy.} 
    \label{figure:fig1}
\end{figure}

Firstly, most mapping models based on the RGB color space perform suboptimally in real-world foggy conditions because fog blurs textures within RGB features, as shown in Fig. \ref{figure:fig1}(a). For instance, MB-Taylor \cite{mb-taylor} (Fig. \ref{figure:fig1}(b)) attempts to extract features and remove degradation in the RGB space but still leaves significant residual haze. However, we have observed that YCbCr features are less affected by fog and provide clearer image textures (Fig. \ref{figure:fig1}(a)). This is because the chrominance components of YCbCr are less responsive to the neutral colors impacted by haze, allowing YCbCr to preserve more color and details in hazy conditions \cite{variational}. Despite their advantages over RGB, converting YCbCr to RGB can introduce rounding errors that cause severe color distortion, as seen with AIPNet \cite{aipnet} in Fig. \ref{figure:fig1}(b). This observation motivates us to use YCbCr features to guide the recovery of clearer structures in RGB mapping models.

\begin{figure}[!t]
    \centering 
    \includegraphics[width=0.97\linewidth]{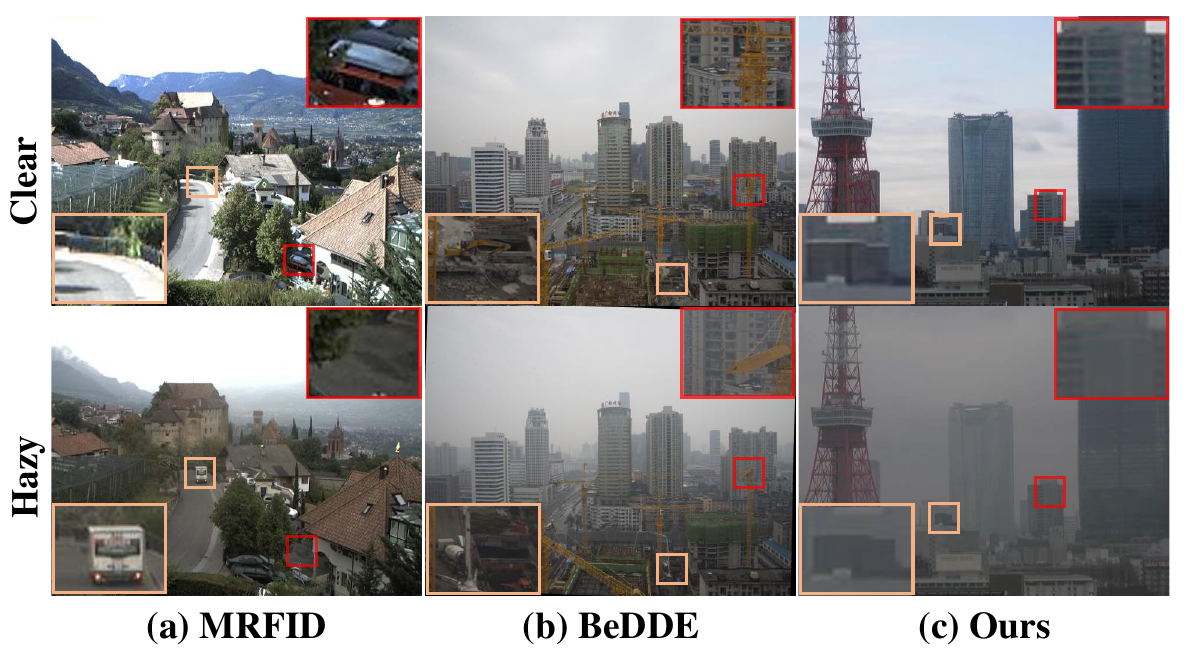}

    \caption{Examples from the MRFID and BEDDE datasets show noticeable background differences between reference and haze images due to varying shooting angles and long time intervals. In contrast, our RW$^2$AH dataset achieves excellent alignment in physical space. }
    \label{figure:fig2}
\end{figure}

Secondly, most existing datasets are unsuitable for fully supervised training (see Fig. \ref{figure:fig2}) due to background inconsistencies \cite{mrfid} and spatial misalignments between hazy/clean images \cite{bedde}. Capturing well-aligned hazy/clean image pairs in real-world conditions is particularly challenging, as uncontrolled environments make it difficult to photograph images with consistent backgrounds within a short timeframe at the same location. Consequently, most dehazing methods are trained on synthetic datasets, which have significant domain gaps from real-world scenes, leading to suboptimal performance in practical applications.

To this end, we propose a novel image dehazing method under RGB and YCbCr color spaces for real haze removal, named the \textbf{S}tructure \textbf{G}uided \textbf{D}ehazing \textbf{N}etwork (SGDN). Specifically, our SGDN employs an asymmetric encoder-decoder structure that shares a single encoder for feature extraction from both RGB and YCbCr color spaces. 
First, we propose a Bi-Color Guidance Bridge (BGB) that enhances structural integrity and refine texture details to improve image quality. It includes two key designs: a Phase Integration Module (PIM) and an Interactive Attention Module (IAM). 
The former guides RGB features to reconstruct clearer textures by the phase spectrum of YCbCr features in the frequency domain, and the latter further fuses the discriminative features of YCbCr space in the spatial domain.
Subsequently, we propose a Color Enhancement Module (CEM) that aggregates channel information from YCbCr features to further enhance the color perception of RGB features.
Finally, we collect a large-scale Real-World Well-Aligned Haze (RW$^2$AH) dataset  that includes multiple scenes. Notably, it can be used for supervised network training and serves as a fair benchmark for evaluating dehazing methods.
Our contributions are summarized as follows:
\begin{itemize}
    \item We propose a novel dehazing method, called SGDN, which exploits the clear texture of the YCbCr to guide the RGB features for real-world image dehazing.

    \item Bi-color Guidance Bridge is proposed to guide the recovery of RGB features from the frequency and spatial domains. Besides, we design the Color Enhancement Module (CEM) to further enhance the color perception capabilities of RGB features.
    
    \item We introduce RW$^2$AH, the first well-aligned real-world dehazing dataset, which contains 1,758 real-world image pairs of corresponding clear and hazy images. While these image pairs may not be perfectly aligned, they are sufficiently aligned for use in supervised learning. This dataset sets a new benchmark for evaluating dehazing methods in real-world image dehazing tasks.
\end{itemize}


\begin{figure*}[!t]
    \centering 
    \includegraphics[width=0.97\linewidth]{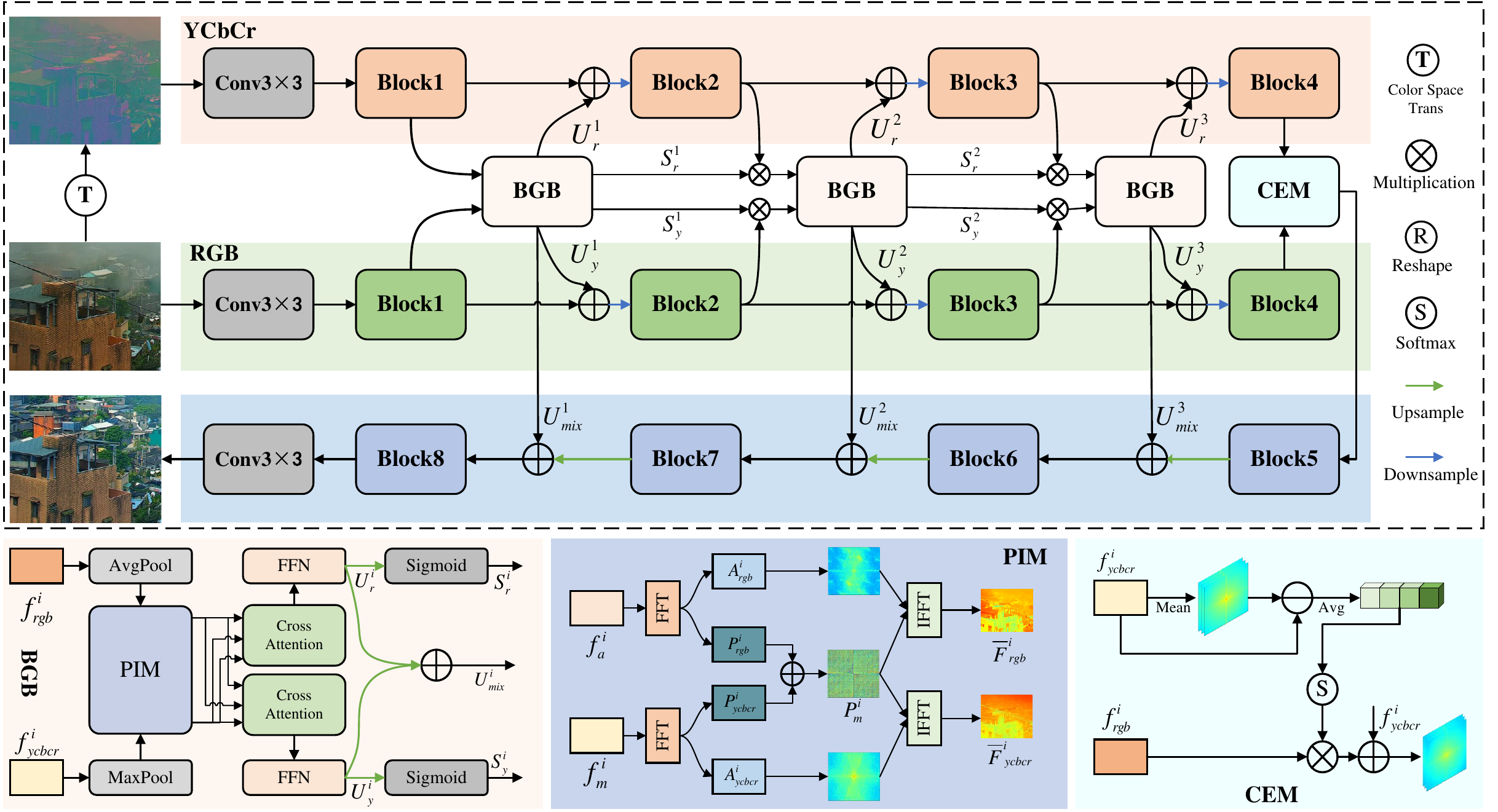}
    \caption{The overall pipeline of our SGDN. It includes the proposed Bi-Color Guidance Bridge (BGB) and Color Enhancement Module (CEM). BGB promotes RGB features to produce clearer textures through YCbCr color space in both frequency and spatial domain, while CEM significantly enhances the visual contrast of the images.}
    \label{figure:framework}
\end{figure*}

\section{Related Works}
\label{relate work}
\subsection{Image Dehazing}
Image dehazing can be regarded as an inverse problem, where the goal is to recover the clean image from a hazy image. 
Here, we primarily review two types of methods:\\
\textbf{Prior-based dehazing methods} \cite{dcp, scep, colorlineprior, yoly} mainly rely on ASM and estimate physical parameters through manual priors or statistical learning methods. For example, the Color Attenuation Prior \cite{cap} estimated the transmittance by analyzing the relationship between color and brightness in foggy images and establishing a linear model. 
Although these methods were effective in some scenarios, they rely on the accuracy of estimating physical parameters, which may produce disappointing results in some more complex real-world scenarios.\\
\textbf{Learning-based dehazing methods} \cite{CDD-GAN, c2pnet, mb-taylor, diacnpn, oknet, DEANet} were proposed to directly estimate haze-free images from hazy inputs.
Recently, some works \cite{nsdnet, fan2024driving, fan2025DCL} have been designed for real-world unpaired dehazing models. 
Some typical examples are built on the CycleGAN \cite{cyclegan} framework, such as DAD \cite{dad}, CycleDehaze \cite{cycledehaze} and D$^4$ \cite{d4}. However, the generative approach of GAN networks often lacks strong constraints, leading to the creation of artifacts that interfere with training. 
Another approaches \cite{psd, refinednet} were to introduce prior knowledge to guide the model's optimization direction. 
For instance, PSD \cite{psd} designed a series of prior losses and fine-tuned real-world dehazing in an unsupervised manner. 
However, these methods cannot avoid the inherent flaws of handcrafted priors. In this paper, we leverage the advantages of dual color spaces to guide the model to learn haze-independent features, achieving superior dehazing results.

\subsection{Color Model in Image Dehazing}
In addition to the commonly used RGB color model, some studies \cite{ryfnet, aipnet, mcpnet, variational} focus on using other color models for haze removal, such as HSV, YCbCr and YUV. 
For example, AIPNet \cite{aipnet} employed a multi-scale network to enhance the contrast of the Y channel, achieving effective haze removal. 
However, these methods often fail when dealing with complex structures due to their reliance on a single assumption. 
To introduce robust constraints, MPCNet \cite{mcpnet} leveraged HSV and YCbCr color spaces, proposing a color correction network and a prior-based dehazing network to improve the removal of non-uniform smoke. 
Although these methods have achieved excellent performance on training data, they require the assistance of ASM and color conversion processes, which inevitably introduce color errors. 

Unlike previous methods, we reveal that YCbCr exhibits clearer features and colors under haze compared to RGB. To avoid conversion errors, we apply YCbCr to guide the reconstruction of haze-independent features in RGB, leading to more effective dehazing.

\section{Methodology}
\label{methods}

The overall framework of our SGDN is illustrated in Fig. \ref{figure:framework}, our SGDN is an asymmetric encoder-decoder structure consisting of Large Kernel Attention Blocks \cite{lsknet}. Given a haze image \( I(x) \in \mathcal{R}^{C \times H \times W} \), we first convert it from RGB to YCbCr space using a linear transformation defined by standard conversion matrices \cite{aipnet}, to obtain \( Y(x) \in \mathcal{R}^{C \times H \times W} \).  Subsequently, we capture the features of different color spaces through the shared encoders of the two branches. Next, we input features from different color spaces into our Bi-color Guidance Bridge, so that RGB features generate the rich refine textures by the YCbCr features. 
Finally, we employ the rich chromaticity information of the YCbCr to enhance the color perception of the RGB, thereby improving the overall visual effect and color expression of the image.

\subsection{Bi-color Guidance Bridge}
In the RGB color space, capturing features with poor background contrast is challenging, but it excels in representing overall image structure and coarse textures. 
Conversely, the YCbCr color space separates luminance and chrominance, enhancing attention to image details and color information. 
Therefore, we propose the BGB to enhance structural integrity and refine texture details by leveraging the inherent advantages of the YCbCr color space. 

Our BGB consist of the a Phase Integration Module (PIM) and an Interaction Attention Module (IAM). As shown in Fig. \ref{figure:framework}, given RGB feature $f_\text{rgb}^{i}\in \mathcal{R}^{C\times H\times W}$, we perform Avgpool operation to smooth out the noise effects, thereby retaining the global structure in a balanced manner. For the YCbCr feature $f_\text{ycbcr}^{i}\in \mathcal{R}^{C\times H\times W}$, we apply Maxpool operation to ensure that image details and colors have higher activation values, making them more robust to minor variations in position and shape. So, they are defined as:
\begin{equation}
    {f}_{a}^{i}, {f}_{m}^{i} = \text{Avgpool}(f_\text{rgb}^{i}), \text{Maxpool}(f_\text{ycbcr}^{i}). 
\end{equation}

\textbf{Phase Integration Module.} Follow the \cite{fsdgn}, clean and hazy images exhibit only minor differences in their phase spectrum. Furthermore, phase spectrum conveys more structural details of the image compared to amplitude spectrum, and it is more robust to contrast distortion and noise. Therefore, we first apply the Fast Fourier Transform (FFT) to convert the features into the frequency domain, and then we can separate the phase and amplitude components:
\begin{equation}
\mathcal{A}_\text{rgb}^{i}, \mathcal{P}_\text{rgb}^{i} = \mathcal{S}(\mathbf{F}(f_{a}^{i})), \ \ \ \mathcal{A}_\text{ycbcr}^{i}, \mathcal{P}_\text{ycbcr}^{i} = \mathcal{S}(\mathbf{F}(f_{m}^{i})), 
\end{equation}
where $\mathcal{A}_\text{rgb}^{i}$ and $ \mathcal{P}_\text{rgb}^{i}$ denote the amplitude and phase spectrum of the RGB feature, respectively. $\mathcal{A}_\text{ycbcr}^{i}$ and $ \mathcal{P}_\text{ycbcr}^{i}$ represent the corresponding features in the YCbCr space.
$\mathcal{S}(\cdot)$ and $\mathbf{F}(\cdot)$ refer to the decoupling and FFT operations.

Considering that amplitude spectrum can suffer from distortion, we employ a $1 \times 1$ convolutional layer to restore the amplitude, and a $3 \times 3$ convolutional layer to extract more detailed structure information. Next, $ \mathcal{P}_\text{rgb}^{i}$ and $ \mathcal{P}_\text{ycbcr}^{i}$ is combined linearly to introduce additional structural details, and obtain a blended phase spectrum $\mathcal{P}_{m}^{i}$. Finally, we perform the inverse Fourier transform using the $\mathcal{P}_{m}^{i}$ and the corresponding amplitude spectrum to reconstruct the features in the spatial domain. This process can be expressed as:
\begin{equation}
\begin{split}
    &\mathcal{P}_{m}^{i} = C_{3\times 3}(\mathcal{P}_\text{ycbcr}^{i}) + C_{3\times 3}(\mathcal{P}_\text{rgb}^{i}), \\
     &\overline{F}_\text{rgb}^{i} = \overline{\mathbf{F}}(C_{1\times 1}(\mathcal{A}_\text{rgb}^{i}), \mathcal{P}_{m}^{i}), \\
    &\overline{F}_\text{ycbcr}^{i} = \overline{\mathbf{F}}(C_{1\times 1}(\mathcal{A}_\text{ycbcr}^{i}), \mathcal{P}_{m}^{i}), 
\end{split}
\end{equation}
where $\overline{F}_\text{ycbcr}^{i}$ and $\overline{F}_\text{ycbcr}^{i}$ are the output of PIM, and $\overline{\mathbf{F}}(\cdot)$ is the inverse Fourier transform. 

\textbf{Interaction Attention Module.} 
Afterwards, IAM is presented to learn important regions of Ycbcr features in the spatial domain. Specifically, IAM utilizes a Cross-Attention (CA) mechanism and a Feed-Forward Network (FFN). For both RGB and YCbCr features, the original feature serves as the query, while the key and value are derived from the features of the other color space. This can be defined as:
\begin{equation}
\begin{split}
    &Q_{r}, K_{r}, V_{r} = \text{Linear}(\overline{F}_\text{rgb}^{i}), \\
    &Q_{y}, K_{y}, V_{y} = \text{Linear}(\overline{F}_\text{ycbcr}^{i}),\\
    &\widetilde{f}_\text{rgb}^{i} = \mathcal{F}_{n}(\mathcal{F}_{c}(Q_{r}, K_{y}, V_{y})),\\ 
    &\widetilde{f}_\text{ycbcr}^{i} = \mathcal{F}_{n}(\mathcal{F}_{c}(Q_{y}, K_{r}, V_{r})),
\end{split}
\end{equation}
where $\text{Linear}(\cdot)$ represents the linear layer and $\mathcal{F}_{n}$ is the FFN, and $\mathcal{F}_{c}$ is the cross attention. 

To further enhance the information flow within the pipeline, we first upsample $\widetilde{f}_\text{rgb}^{i}$ and $\widetilde{f}_\text{ycbcr}^{i}$ to obtain $U_{r}^{i}$ and $U_{y}^{i}$. Next, we compress them to the range (0,1) to obtain $S_{r}^{i}$ and $S_{y}^{i}$, and propagate them effectively to the next stage through matrix multiplication. This ensures the maximization of both the global structure and image details during the fusion process. Finally, we perform element-wise addition operation on $U_{r}^{i}$ and $U_{y}^{i}$ to obtain $U_{mix}^{i}$ as the residual of the decoder. It can be represented by the following formula:
\begin{equation}
\begin{split}
    & U_{r}^{i}, \ U_{y}^{i}  = Up(\widetilde{f}_\text{rgb}^{i}), \ Up(\widetilde{f}_\text{ycbcr}^{i}), \\
    &f_\text{rgb}^{i+1}, \ f_\text{ycbcr}^{i+1} = \mathbf{Sig}(U_{r}^{i}) \otimes f_\text{rgb}^{i+1}, \ \mathbf{Sig}(U_{y}^{i}) \otimes f_{y}^{i+1}, \\
    &U_{mix}^{i} = U_{r}^{i} + U_{y}^{i},
\end{split}
\end{equation}
where $\mathbf{Sig}(\cdot)$ represents the sigmoid operation and $\otimes$ denotes element-wise multiplication.

\subsection{Color Enhancement Module}
In the YCbCr color space, the chrominance components have a constant or zero response to neutral colors directly affected by haze, while the luminance is proportional to the haze density. This allows YCbCr to retain more colors and image details in hazy environments. Therefore, we propose a Color Enhancement Module (CEM) that leverages the rich color information in the YCbCr color space to enhance the RGB features. Formally, let $f_\text{rgb}^{i}$ and $f_\text{ycbcr}^{i}$ denote the inputs of the CEM. We further remove global illumination and haze effects from $f_\text{ycbcr}^{i}$ to emphasize local variations and color. We achieve this by averaging across channels at each position and subtracting this mean, centering the pixel values symmetrically around zero:
\begin{equation}
    \mathbf{A}_{m}=f_\text{ycbcr}^{i} - \mathcal{M}(f_\text{ycbcr}^{i}),
\end{equation}
where $\mathbf{A}_{m}$ is local concentration features, and $\mathcal{M}(\cdot)$ is the mean normalization. We then transform $\mathbf{A}_{m}$ into distribution intensity $\mathbf{v}_{c}$ using a project network with global average pooling and softmax. Finally, $\mathbf{v}_{c}$ modulates $f_\text{rgb}^{i}$ to enhance local RGB feature perception while preserving the gradient in $f_\text{ycbcr}^{i}$. This can be defined as:
\begin{equation}
    \mathbf{v}_{c}=\text{Softmax}(\text{Avg}(\mathbf{A}_{m})), \ \
    \mathbf{D}_{o}=\mathbf{v}_{c} \otimes f_\text{rgb}^{i} + f_\text{ycbcr}^{i},
\end{equation}
where $\mathbf{D}_{o}$ denotes the final output of CEM. 
\subsection{Training Loss}
 Follow the \cite{fsnet}, we employ multi-scale loss to train our SGDN. The overall loss can be expressed as:
\begin{equation}
    \mathcal{L}_{all} = \sum_{s=1}^{3} \lbrack \eta \mathcal{L}_{\ell_1}^{s} + \theta \mathcal{L}_\text{ssim}^{s} + \lambda \mathcal{L}_\text{fft}^{s} \rbrack ,
\end{equation}
where $\mathcal{L}_{\ell_1}$ represents $\ell_1$ loss, $\mathcal{L}_\text{ssim}$ is the structural similarity loss \cite{ssimloss}, and $\mathcal{L}_\text{fft}$ means Fourier loss \cite{fftloss}; $s$ is the input and output of different sizes, with values of 1, 0.5, and 0.25.
$\eta, \theta$ and $\lambda$ represent the weight coefficient of the corresponding loss function, which are set to 1.0, 0.5, and 0.1, respectively. In addition, $\mathcal{L}_\text{ssim}$ can better preserve the details and color information of the image in cases of poor alignment, making the resulting image look more natural. \\

\section{Real-World Well-Aligned Haze Dataset}
\label{RW$^2$AHD}
\begin{figure}[!t]
    \centering 
    \includegraphics[width=0.97\linewidth]{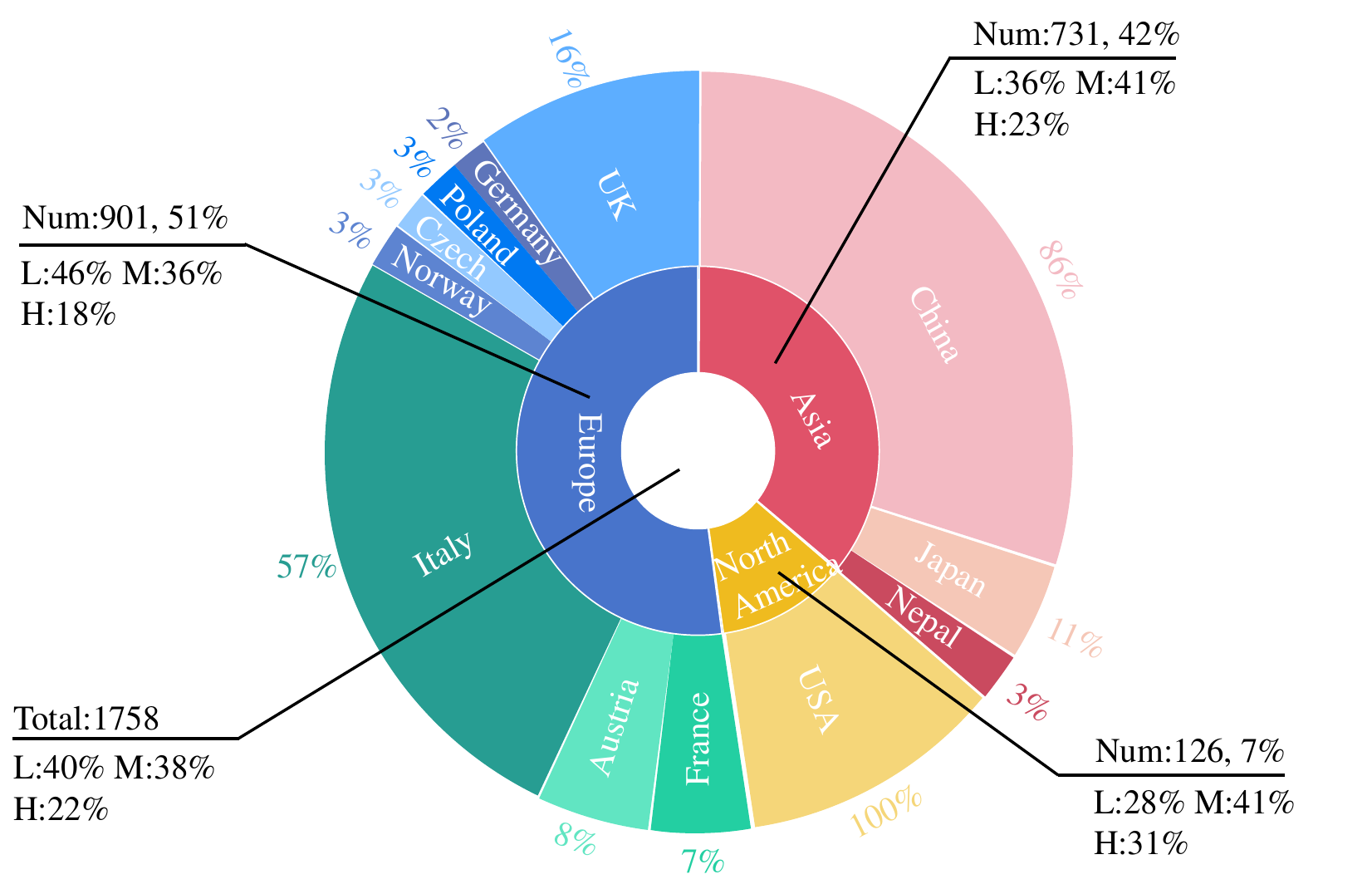}
    \caption{Geographic and haze distribution of our RW$^2$AH.}
    \label{figure:fig4}
\end{figure}

To address the lack of paired real-world haze datasets for fully supervised learning, we create a Real-world Well-Aligned Haze (RW$^2$AH) dataset comprising 1,758 pairs. The collection process and statistical analysis are as follows:

\begin{itemize}
    \item 1) \textbf{Source Selection}: We use stationary webcams from YouTube to capture hazy and clean images, spanning diverse environments such as landscapes, vegetation, buildings, and mountains.
    \item 2) \textbf{Geographical Coverage}: The dataset includes webcams from twelve countries across Asia, Europe, and America, capturing various climate conditions and haze formations.
    \item 3) \textbf{Temporal Pairing}: Hazy and clean images are recorded on the same day, ensuring they are captured from the same webcam, with intervals ranging from 1 to several hours. 
    \item 4) \textbf{Haze Density}: We categorize the captured haze images into three density levels: light haze (40\%), moderate haze (38\%), and heavy haze (22\%), as shown in Fig. \ref{figure:fig4}.
    \item 5) \textbf{Exclusion of Dynamic Objects}: To maintain scene consistency, we exclude images with uncontrollable objects (e.g., vehicles, pedestrians) and those where more than two-thirds of the image is sky.
    \item 6) \textbf{Cropping}:  To ensure background alignment, we apply artificial cropping, keeping the most similar image pairs and minimizing discrepancies from slight camera shifts or lighting changes. 
\end{itemize}





Compared to the BeDDE (208 pairs) \cite{bedde} and the MRFID (800 pairs) \cite{mrfid} datasets, our RW$^2$AH (1,758 pairs) offers a larger volume of data and greater scene diversity. Notably, existing supervised models trained on our dataset using only $\ell_1$ loss achieve excellent performance. As a result, our RW$^2$AH can serve as a new benchmark for evaluating real-world dehazing methods using standard metrics like PSNR and SSIM.


\begin{figure*}[!t]
    \centering 
    \includegraphics[width=0.97\linewidth]{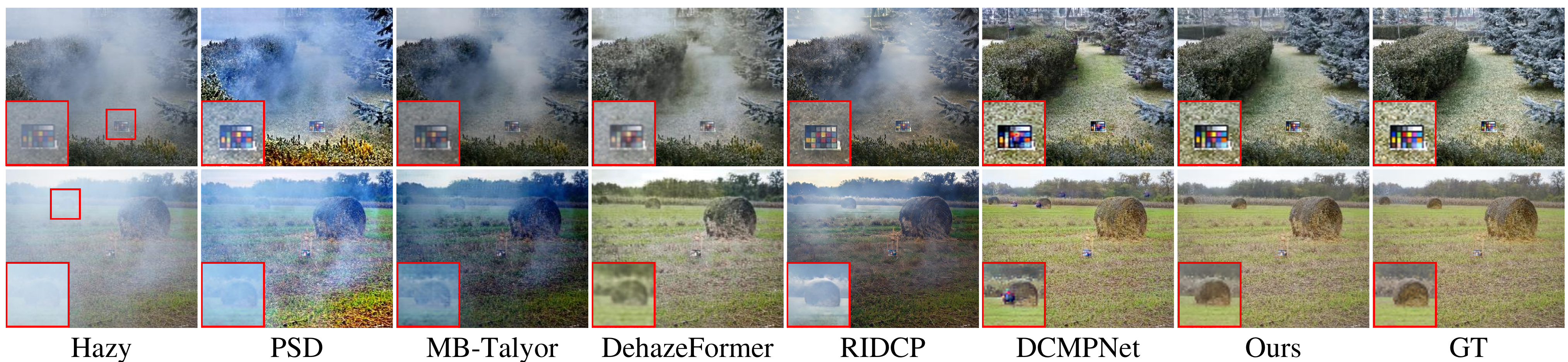}

    \caption{Visual comparison results on the real-world smoke dataset. Zoom in for a better view.} 
    \label{figure:fig5}
\end{figure*}

\begin{figure*}[!t]
    \centering 
    \includegraphics[width=0.97\linewidth]{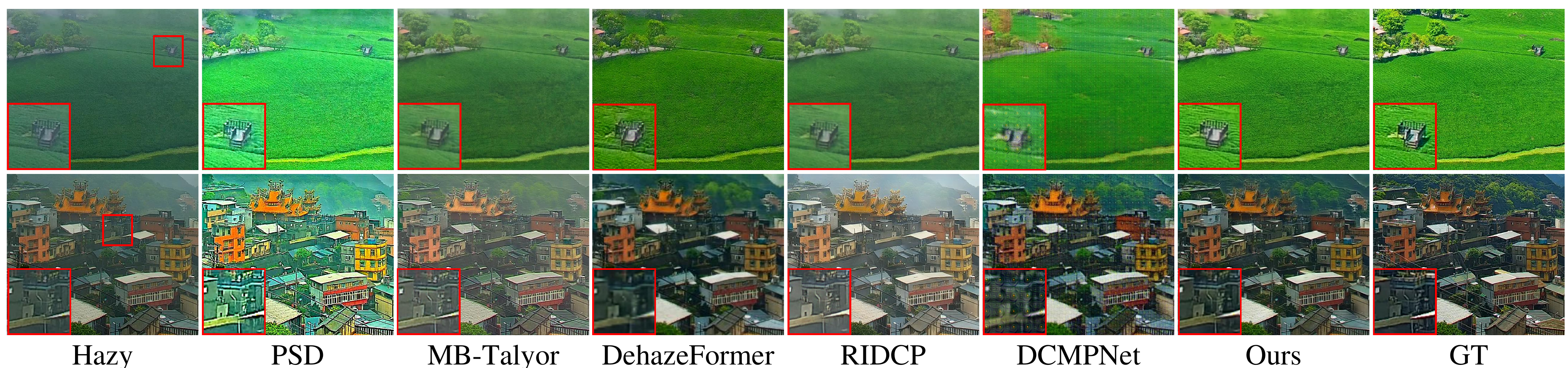}
    \caption{Visual comparison results on the our RW$^2$AH dataset. Zoom in for a better view.}
    \label{figure:fig6}
\end{figure*}


\begin{table*}[!t]
    \centering
    \renewcommand{\arraystretch}{1.2}
    \scalebox{0.72}{
    \begin{tabular}{l|c|c|cccc|cccc|cc|cc}
    \hline
    \Xhline{1.2pt}
          \multirow{2}{*}{\textbf{Data Settings}} & \multirow{2}{*}{\textbf{Methods}} & \multirow{2}{*}{\textbf{Venue}} & \multicolumn{4}{c|}{\textbf{Real-world Smoke}} & \multicolumn{4}{c|}{\textbf{RW$^2$AH (Ours)}}& \multicolumn{2}{c|}{\textbf{RTTS}} & \multicolumn{2}{c}{\textbf{Overhead}} \\ 
        
         ~ & ~ & ~ & PSNR$\uparrow$& SSIM$\uparrow$ & FADE$\downarrow$ & NIQE$\downarrow$ & PSNR$\uparrow$ & SSIM$\uparrow$ & FADE$\downarrow$ & NIQE$\downarrow$ & FADE$\downarrow$ & NIQE$\downarrow$ & Params & FLOPS \\ \hline \hline
        
         \multirow{4}{*}{Unpaired} & DCP  & CVPR'09 & 15.01 & 0.392  & 0.4961 & 4.1303 & 15.56 & 0.479 & 0.5716 & 5.8557 & 0.7865 & 6.6401 & - & -  \\
         
        
          ~ & RefineDNet & TIP'21 & 17.44 & 0.525  & 0.6042 & 5.1060 & 16.62 & 0.517 & 0.4908 & 5.8190 & 0.6149 & 6.9708 & 65.78M & 276.81G   \\ 

         ~ & CDD-GAN &  ECCV'22 & 12.16 & 0.250  & 0.3174 & 5.3888 & 17.78 & 0.601 & \underline{0.4111} & 5.7625 & 0.7470 & 6.6245 & 29.27M & 279.44G  \\
        
          ~ & D$^{4}$ & CVPR'22 & 18.21 & 0.644  & 0.6907 & 5.0909 & 15.68 & 0.519 & 0.6791 & 7.8726 & 0.7444 & 5.7960 & 10.7M & \textbf{2.25G} \\ \hline

         \multirow{12}{*}{Paired}  & PSD & CVPR'21 & 12.86 & 0.440  & 1.1304 & 4.4485 & 16.95 & 0.448 & 0.6204 & 5.9418 & 0.6115 & 6.0018 & \underline{6.21M} & 143.91G \\ 
        
        ~ & YOLY & IJCV'21 & 14.29 & 0.409  & 0.9502 & 4.9007 &  15.40 & 0.451 & 0.4827 & 6.0521 & 0.8663 & 7.0794 & \textbf{1.24M} & 303.90G  \\ 
        
        ~ & FSDGN & ECCV'22 & 19.43 & 0.671  & 0.9904 & 5.7410 & 18.57 & 0.430 & 0.8973 & 6.4823 & 0.8034 & 6.7189 & \underline{2.73M} & \underline{19.6G} \\ 
        
        ~ & DehazeFormer & TIP'23 & 18.64 & 0.688  & 0.4580 & 4.1083 & 20.36 & 0.612 & 0.6904 & 5.8067 & 0.6151 & 5.8130 & 25.44M & 139.85G  \\

       ~ & C$^2$PNet & CVPR'23 & 17.86 & 0.361  & 0.8795 & 6.1102 & 17.48 & 0.472 & 0.6154 & 7.0129 & 0.6086 & 5.7815 & 7.17M & 460.95G  \\ 

       ~ & RIDCP & CVPR'23 & 19.86 & 0.618 & 0.3932 & 3.7959 & 19.87 & \underline{0.627} & 0.5147 & 5.3657 & \underline{0.4796} & \textbf{5.0843} & 28.72M & 188.65G   \\ 
        
        ~ & MB-Talyor & ICCV'23 & 16.49  & 0.616  & 0.6911 & 6.1694 & 19.49 & 0.609 & 0.7979 & 5.9100 & 0.5680 & 6.9914 & 7.43M & 44.05G  \\ 

         ~ & NSDNet & arXiv'23 & 19.92 & \underline{0.780}  & \underline{0.3031} & 3.7884 & \underline{21.39} & 0.619 & 0.4576 & \underline{5.3337} & 0.4821 & 5.3629 & 11.38M & 56.86G \\ 
         
        ~ & DAE-Net & TIP'24 & \underline{22.59}  & 0.778  & 0.5524 & 3.9841 & 21.14 & 0.574 & 0.8042 & 5.4397 & 0.7297 & \underline{5.1751} & 3.65M & \underline{32.23G}  \\ 

        ~ & OKNet & AAAI'24 & 21.93 & 0.750  & 0.3379 & \underline{3.5567} & 21.28 & \underline{0.627} & 0.7124 & 6.0040 & 0.7512 & 6.0741 & 4.72M & 39.67G  \\ 
        
        ~ & DCMPNet & CVPR'24 & 21.01  & 0.767 & 0.3740 & 3.8295 & 20.13 & 0.587 & 0.4726 & 5.6546 & 0.6257 & 8.4573 & 7.16M & 62.89G   \\ \cline{2-15}
        
         ~ &  Ours & - & \textbf{23.41} & \textbf{0.790}  & \textbf{0.3042} & \textbf{3.4365} & \textbf{22.26} & \textbf{0.668} & \textbf{0.4001} & \textbf{5.0080} & \textbf{0.4611} & 5.2114 & 13.32M & 53.40G  \\ 
         \Xhline{1.2pt}
          \hline
    \end{tabular}
    }
    \caption{Quantitative study on real-world smoke/haze datasets, $\downarrow$ indicates that lower values are better, while $\uparrow$ indicates that higher values are better. The best results are highlighted in bold, and the second-best results are underlined.}
    \label{table:table1}
\end{table*}

\begin{figure*}[!t]
    \centering 
    \includegraphics[width=0.97\textwidth]{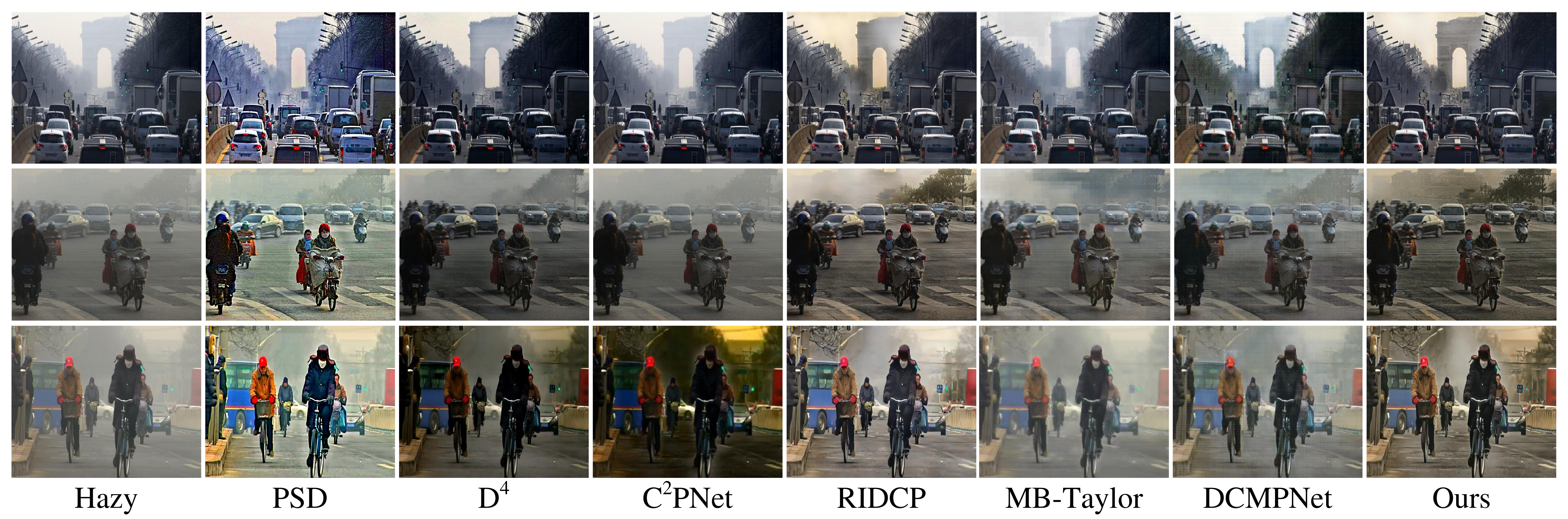}
    \caption{Visual comparison results on the RTTS dataset.}
    
    \label{figure:fig7}
\end{figure*}

\section{Experiments}
\label{experiment}
\subsection{Datasets and Metrics}
We conduct qualitative and quantitative studies on three real-world smoke/haze datasets. Here, we employ four metrics to evaluate the dehazing effectiveness and image quality. These include the commonly used reference quality assessment metrics, PSNR and SSIM. Additionally, we employ the no-reference image evaluation metrics, FADE \cite{fade} and NIQE \cite{niqe}, to measure the haze density and image quality.

\textbf{Our RW$^2$AH dataset} is a real-world haze dataset collected from global online webcams, encompassing various climate conditions and diverse scenes from different geographic locations. The dataset includes 1,406 pairs of images with different resolutions for training and 352 pairs of images for evaluation. It features well-aligned backgrounds, making it suitable for effective supervised learning.

\textbf{Real-World Smoke (RWS) Dataset} is a collection of I-HAZE \cite{ihaze}, O-Haze \cite{o-haze}, and NH-Haze \cite{nhhaze} datasets. It contains 155 pairs of homogeneous and non-homogeneous indoor and outdoor smoke images. Following \cite{nsdnet}, we select 147 images pairs for training and 8 images pairs for evaluation.

\textbf{Real-world Task-Driven Testing Set (RTTS)} is a subset of the RESIDE \cite{rtts} dataset, comprising 4,322 outdoor real-world haze images. Since it lacks corresponding ground truth, it is only used for testing purposes.

\subsection{Comparison with State-of-the-arts}
To validate the effectiveness of our SGDN in real haze/ smoke scenarios, we compare it with several representative SOTA methods. For fairness, we use their default settings and fine-tune them on our RW$^2$AH dataset for optimal performance. Table \ref{table:table1} reports the quantitative comparison results on three real-world haze/smoke datasets.  

\textbf{Results on RWS.} As depicted in Table \ref{table:table1}, our SGDN outperforms all SOTA methods. Specifically, it surpasses DCMPNet \cite{diacnpn} by approximately 10.99\% in PSNR, 2.99\% in SSIM, 18.66\% in FADE, and 9.27\% in NIQE. Additionally, Fig. \ref{figure:fig5} displays the visual results of each method on real smoke. PSD \cite{psd}, MB-Taylor \cite{mb-taylor}, DehazeFormer \cite{dehazeformer}, and RIDCP \cite{ridcp} typically struggle with dense smoke, as they all learn smoke mapping in the RGB color space, which makes it difficult to capture useful textures among blurry features. DCMPNet, with robust depth estimation, performs better in dense smoke but still shows minor color deviations. In contrast, our SGDN uses YCbCr texture details to guide RGB dehazing, producing results closer to the clear ground truth.

\textbf{Results on our RW$^2$AH.} Fig. \ref{figure:fig6} illustrates the dehazing comparison of various methods on our RW$^2$AH dataset. Our method outperforms the others in dehazing effectiveness. Specifically, PSD significantly enhances image contrast but fails to fully eliminate haze. MB-Taylor and RIDCP still retain a considerable amount of haze in their results. Although DehazeFormer and DCMPNet remove haze effectively, they struggle to preserve image details and textures. In contrast, our method achieves better dehazing while maintaining the natural contrast. Additionally, our method demonstrates stronger performance in the quantitative comparisons reported in Table \ref{table:table1}.

\textbf{Results on RTTS.} From Table \ref{table:table1}, our SGDN achieves the highest FADE score, indicating its robust dehazing capability. In terms of NIQE, SGDN delivers a competitive result, slightly trailing the top-performing RIDCP, while maintaining strong overall visual quality. Additionally, Fig. \ref{figure:fig7} shows that most methods struggle to effectively remove haze, particularly in dense fog conditions, while our approach significantly outperforms the others.

\begin{figure}[!t]
    \centering 
    \includegraphics[width=0.97\linewidth]{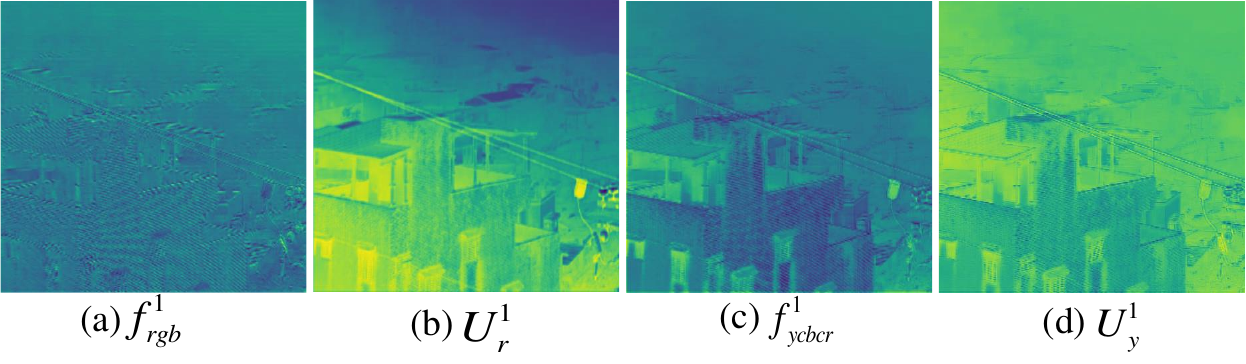}
    \caption{Visualization of BGB inputs ($f_\text{rgb}^{1}$ and $f_\text{ycbcr}^{1}$) and outputs ($U_{r}^{1}$ and $U_{y}^{1}$). }
    \label{figure:feature_bcb}
\end{figure}

\begin{table}[t]
	\linespread{1.0}
	\centering
	\setlength\tabcolsep{1.8pt} 
	\renewcommand\arraystretch{1.2}
	\scalebox{0.83}{	
		\begin{tabular}{l|cc|cccc}
			\Xhline{1.3pt} 
			\multicolumn{1}{c|}{\multirow{2}[1]{*}{\makecell{Ablations for\\color space}}} & \multicolumn{2}{c|}{Real-world Smoke} & \multicolumn{4}{c}{Our RW$^2$AH} \\
			& \multicolumn{1}{c}{PSNR$\uparrow$} & \multicolumn{1}{c|}{SSIM$\uparrow$} & \multicolumn{1}{c}{PSNR$\uparrow$} & \multicolumn{1}{c}{SSIM$\uparrow$}& \multicolumn{1}{c}{FADE$\downarrow$} & \multicolumn{1}{c}{NIQE$\downarrow$}\\
			\Xhline{0.7pt} 
			Only RGB & 19.36 & 0.593 & 18.44 & 0.417 & 0.679 & 7.190 \\
			Only YCbCr & 19.48 & 0.607 & 19.02 & 0.472 & 0.640 & 6.971 \\
			RGB+HSV & 22.46   & 0.703   & 21.17  & 0.593  & 0.556 & 6.748 \\
            RGB+YUV & 23.04   & 0.736   & 21.49  & 0.608  & 0.489 & 5.878 \\
			\rowcolor{gray!10} Ours & \textbf{23.41} & \textbf{0.790} & \textbf{22.26} & \textbf{0.668} & \textbf{0.400} & \textbf{5.008} \\  
			\Xhline{1.3pt} 
	\end{tabular}}

	\caption{Quantitative study of different color models on our RW$^2$AH dataset and real-world smoke dataset.}
	\label{table2:ablation study}
\end{table}

\begin{figure}[t]
	\centering
	\includegraphics[width=0.97\linewidth]{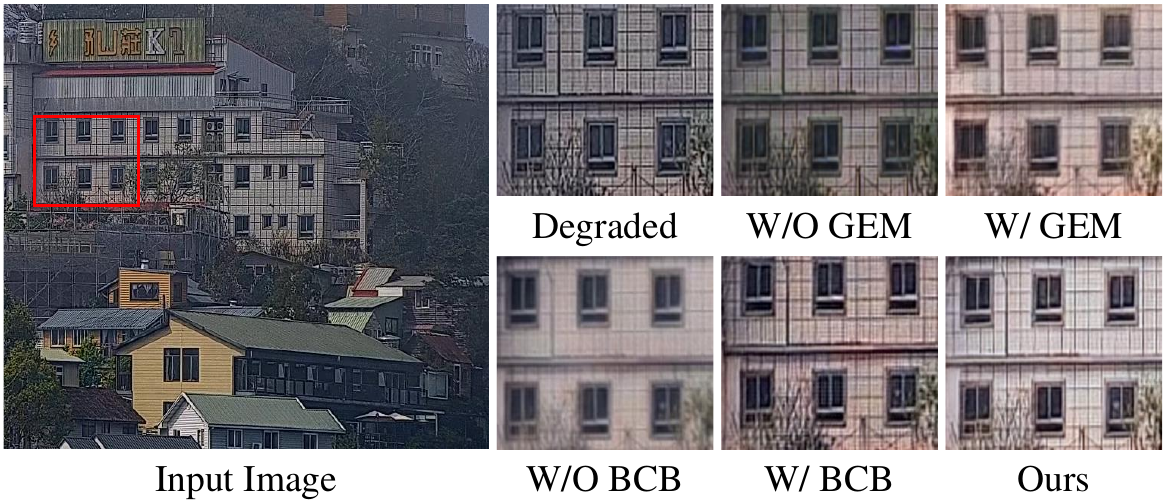}
	\caption{Ablation study of CEM. Our CEM can significantly improve the visual contrast of the image.}
	\label{fig:8}
\end{figure}

\begin{table}[t]
	\linespread{1.0}
	\centering
	\setlength\tabcolsep{1.8pt} 
	\renewcommand\arraystretch{1.2}
	\scalebox{0.83}{	
		\begin{tabular}{l|cc|cccc}
			\Xhline{1.3pt} 
			\multicolumn{1}{c|}{\multirow{2}[1]{*}{\makecell{Ablations for\\BGB and CEM}}} & \multicolumn{2}{c|}{Real-world Smoke} & \multicolumn{4}{c}{Our RW$^2$AH} \\
			& \multicolumn{1}{c}{PSNR$\uparrow$} & \multicolumn{1}{c|}{SSIM$\uparrow$} & \multicolumn{1}{c}{PSNR$\uparrow$} & \multicolumn{1}{c}{SSIM$\uparrow$}& \multicolumn{1}{c}{FADE$\downarrow$} & \multicolumn{1}{c}{NIQE$\downarrow$}\\
			\Xhline{0.7pt} 
			Baseline & 20.17 & 0.603 & 19.10 & 0.468 & 0.613 & 7.003 \\
			+BGB & 22.75   & 0.686  & 21.83  & 0.641  & 0.497 & 6.077 \\
			+CEM & 21.46   & 0.627  & 21.36  & 0.605  & 0.607 & 6.917 \\
			\rowcolor{gray!10} +BGB+CEM & \textbf{23.41} & \textbf{0.790} & \textbf{22.26} & \textbf{0.668} & \textbf{0.400} & \textbf{5.008}\\  
			\Xhline{1.3pt} 
	\end{tabular}}
	\caption{Ablation study of our proposed module.}
	\label{table3:ablation study3}
\end{table}

\subsection{Ablation Study}
Our proposed SGDN primarily incorporates the YCbCr color model, BGB, and CEM. To evaluate their contributions to the overall performance, we conduct a series of studies on our RW$^2$AH and real-world smoke datasets.\\
\textbf{Effect of different color models.} To verify the beneficial impact of the different color models, we conduct comparative experiments, with results presented in Table \ref{table2:ablation study}.  The findings show that models using only a single color space are limited in performance. In contrast, models combining RGB and HSV color spaces outperform those using either space alone. This is because the hue and saturation channels in HSV lack edge information \cite{ryfnet}, making it difficult to effectively enhance RGB features. In comparison, our SGDN recovers clean features with the help of YCbCr, achieving the best overall performance. Additionally, we visualize the input and output of BGB in Fig. \ref{figure:feature_bcb}. The input $f^{1}\text{rgb}$ of BGB produces more detailed texture in $U^{1}{r}$ when guided by $f^{1}\text{ycbcr}$. Meanwhile, $U^{1}{y}$ offers a more comprehensive global response after BGB processing.\\
\textbf{Effect of BGB and CEM.} Table \ref{table3:ablation study3} reports the ablation results for BGB and CEM. We exclude BGB and CEM, and form a baseline by adding the YCbCr and RGB branches. The results show that BGB and CEM bring gains of 2.73 dB and 2.26 dB on our RW$^2$AH dataset, respectively. The gain from BGB is more significant as it promotes clearer RGB features over Ycbcr features. While the gain from CEM is smaller, it significantly improves visual quality. In Fig. \ref{fig:8}, the model using CEM can restore better colors, bringing the images closer to the ground truth, thus validating its effectiveness. Finally, when both BGB and CEM are combined, our SGDN achieves its best performance.

\section{Conclusion}
\label{conclusion}
In this paper, we propose the \textbf{S}tructure \textbf{G}uided \textbf{D}ehazing \textbf{N}etwork (SGDN), a novel framework specifically designed for the real-world dehazing. It simultaneously extracts features from both RGB and YCbCr color spaces to better collaborate in removing haze. To guide RGB features to produce clearer textures, we design the Bi-Color Guidance Bridge (BGB). Given the more pronounced color information in the YCbCr color space, we propose the Color Enhancement Module (CEM) to enhance the natural color perception in the RGB color space. Due to the effectiveness of our designs, our SGDN achieves optimal performance on several real-world haze datasets. Finally, to advance the development of real-world dehazing task, we construct a new trainable real-world well-aligned (RW$^2$AH) dataset, which includes a diverse range of geographical regions and climate conditions worldwide, ensuring the dataset's diversity.

\section*{Acknowledgements} This work was supported by the National Science Fund of China under Grant Nos. 62072242, U24A20330 and 62361166670.

\bibliography{aaai25}
\newpage

\begin{figure*}[!t]
    \centering 
    \includegraphics[width=1\textwidth]{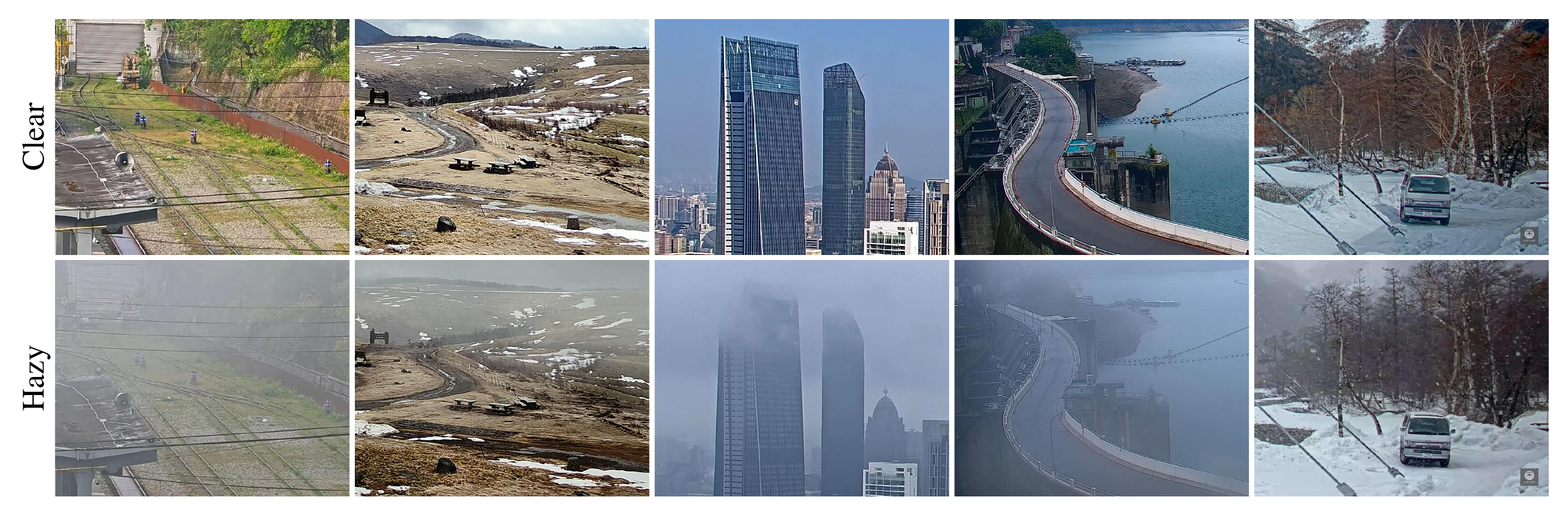}
    \vskip -0.1in
    \caption{Example images on the RW$^2$AH dataset. It has good reference images as supervision signals and covers multiple scenes}
    \label{figure:appendix_dataset}
\end{figure*}

\section{Appendix}
This appendix is organized as follows:
\begin{itemize}
    \item In Section A, we provide more ablation experiments and quantitative comparisons of our method on synthetic datasets.
    \item In Section B, we provide the impact of our approach on the downstream task of image segmentation in a real-world setting.
    \item In Section C,  we provide more visual results on real-world haze conditions.
\end{itemize}

\section{A. More Experiments}
\label{appendix_ablation}
\subsection{Ablation Study of component in BGB.}
\label{appendix_ablation_bcb}

To validate the impact of each component in the proposed BGB on performance, we exclude the interference of CEM and incrementally add each subcomponent to join BGB. As shown in Table \ref{table4:ablation study3}, we can clearly see the the IAM contributes the most, improving the PSNR by 2.5 dB in our RW$^2$AH dataset. While the benefit of PIM is slightly less than that of IAM, it does not incur any additional computational effort and has only 0.32M parameters. But combined with the two, our model has a significant improvement.

\begin{table}[h]
	\linespread{1.0}
	\centering
	\setlength\tabcolsep{1.8pt} 
	\renewcommand\arraystretch{1.3}
	\scalebox{0.83}{	
		\begin{tabular}{l|cccc|cc}
			\Xhline{1.3pt} 
			\multicolumn{1}{c|}{\multirow{2}[1]{*}{\makecell{Component in BGB}}}  & \multicolumn{4}{c|}{Our RW$^2$AH} & \multicolumn{2}{c}{Overall} \\
			& \multicolumn{1}{c}{PSNR$\uparrow$}  & \multicolumn{1}{c}{SSIM$\uparrow$}& \multicolumn{1}{c}{FADE$\downarrow$} & \multicolumn{1}{c|}{NIQE$\downarrow$} & \multicolumn{1}{c}{Params} & \multicolumn{1}{c}{FLOPS}\\
			\Xhline{0.7pt}
            Baseline  & 19.10 & 0.468 & 0.613 & 7.003 & 10.30M & 19.14G\\
			+PIM   & 21.52  & 0.604  & 0.534 & 6.417 & 10.62M   & 19.14G \\
            +IAM   & 21.60  & 0.619  & 0.518 & 6.201 & 13.32M   & 53.40G \\
			\rowcolor{gray!10} +PIM+IAM   & 21.83  & 0.641  & 0.497 & 6.077 & 13.32M   & 53.40G \\
			\Xhline{1.3pt} 
	\end{tabular}}
	\caption{Ablation study of each component in BGB.}
	\label{table4:ablation study3}
\end{table}

\begin{table}[h]
	\linespread{1.0}
	\centering
	\setlength\tabcolsep{1.8pt}
	\renewcommand\arraystretch{1.4}
	\scalebox{1.0}{
		\begin{tabular}{c|cc|cc}
			\Xhline{1.3pt}
			\multicolumn{5}{c}{\makecell{Fix $\mathcal{L}_{\ell_1}:\mathcal{L}_{ssim}:\mathcal{L}_{fft}$ to $10:5:1$}}  \\
			\Xhline{0.7pt}
			~  & \multicolumn{1}{c}{PSNR $\uparrow$} & \multicolumn{1}{c|}{SSIM $\uparrow$} & \multicolumn{1}{c}{FADE $\downarrow$} & \multicolumn{1}{c}{NIQE $\downarrow$}\\
			\Xhline{0.7pt}
			$\mathcal{L}_{\ell_1}$ (Base) & 21.90  & 0.609 & 0.4211 & 5.7789\\
			$\mathcal{L}_{l\ell_{1}+\text{ssim}}$  & 21.93 & 0.627 & 0.4209 & 5.4289 \\
            \rowcolor{gray!10}$\mathcal{L}_{\ell_1+\text{ssim}+\text{fft}}$ & 22.26 & 0.668 & 0.4001 & 5.0080 \\
			\Xhline{1.3pt}
	\end{tabular}}%
	\caption{An ablation study was conducted on our RW$^2$AH dataset to investigate the effects of different loss components.}
	\label{tab5:different loss items}
\end{table}%

\subsection{Ablation Study of different function losses.} 
\label{appendix_ablation_loss}
To investigate the effects of different loss components on our model's performance, we conducted an ablation study using the our RW$^2$AH dataset. The results are presented in Table \ref{tab5:different loss items}. When using \(\mathcal{L}_{\ell_1}\) loss alone, the model achieves a PSNR of 21.90 and an SSIM of 0.609. However, the FADE and NIQE values are relatively high, indicating that the model's reconstructed images have limited quality and clarity. Next, we introduced \(\mathcal{L}_{\text{ssim}}\), resulting in a significant increase in PSNR and SSIM. The FADE and NIQE values also decreased, indicating improvements in image sharpness and overall quality. The increase in SSIM highlights the effectiveness of \(\mathcal{L}_{\text{ssim}}\) in capturing structural information and enhancing image details.
We further incorporated \(\mathcal{L}_{\text{fft}}\) into the total loss, which significantly improved performance. PSNR and SSIM increased by 0.15 and 0.021 respectively, indicating better reconstruction quality. The decrease in FADE and NIQE values reflects enhanced image sharpness and reduced artifacts.

\begin{table}[!t]
	\linespread{1.0}
	\centering
	\setlength\tabcolsep{1.3pt}
	\renewcommand\arraystretch{1.4}
	\scalebox{0.84}{
		\begin{tabular}{l|cc|cc|c|c}
			\Xhline{1.3pt}
			\multicolumn{1}{l|}{\multirow{2}[1]{*}{Methods}}& \multicolumn{2}{c|}{Indoor} & \multicolumn{2}{c|}{Outdoor} &\multirow{2}[1]{*}{\makecell{FLOPS\\(G)}} &\multirow{2}[1]{*}{Reference}\\
			& PSNR $\uparrow$  & SSIM $\uparrow$ & PSNR $\uparrow$  & SSIM $\uparrow$ &  & \\
			\Xhline{0.7pt}
            PSD & 12.50  & 0.715 & 15.51 & 0.748 & 143.91 & CVPR'21\\
            D$^4$ & 25.24  & 0.932 & 25.83 & 0.956 & \textbf{2.25} & CVPR'22\\
			FFA-Net & 36.39  & 0.989 & 33.57 & 0.579 & 287.8 & AAAI'20\\
			SGID & 38.52  & 0.991 & 30.20 & 0.986 & 156.6 & TIP'22\\
			DeHamer & 36.63   & 0.988 & 35.18 & 0.986 & 48.93 & CVPR'20\\
			PMNet& 38.41  & 0.990 & 34.74 & 0.985 & 81.13 & ECCV'20\\
            MAXIM-2S & 38.11   & 0.991  & 34.19  & 0.985 & 216.0 & CVPR'22\\
			DehazeFormer & 38.46  & 0.994 & 34.29 & 0.983 & 47.32 & TIP'23\\
			C$^2$PNet & \textbf{42.56} & 0.995 & \textbf{36.68} & \underline{0.990} & 460.95 & CVPR'23\\
            DEA-Net & 41.31 & 0.994 & \underline{36.59} & 0.989 & \underline{32.23} & TIP'24\\
			DCMPNet & 42.18 & 0.996  & 36.56 & \textbf{0.993} & 62.89 & CVPR'24\\
			\Xhline{0.7pt}
			\rowcolor{gray!10}Ours & \underline{42.29} & \textbf{0.998} & 36.22 & 0.986   & 53.40 & - \\
			\Xhline{1.3pt}
	\end{tabular}}%
	\caption{Dehazing Comparisons on the SOTS-Indoor and SOTS-Outdoor dataset. The 1st and 2nd best results are emphasized with bold and underline, respectively.}
	\label{tab:Appendix_synthetic}%
\end{table}%

\subsection{Experiments on Synthetic Datasets}
\label{appendix_syn_dataset}
Table \ref{tab:Appendix_synthetic} presents a comparison of various dehazing methods on the SOTS-Indoor and SOTS-Outdoor datasets using PSNR and SSIM as evaluation metrics. Our method demonstrates impressive results, particularly in the indoor setting, achieving the second-best PSNR and the best SSIM. It also performs competitively in the outdoor setting.

Specifically, in the indoor scenario, our method outperforms all others in SSIM, indicating superior preservation of structural information in the dehazed images. With a PSNR of 42.29, it is only marginally behind the top-performing C$^2$PNet and DCMPNet. For outdoor images, our method maintains a strong performance with a PSNR of 35.22 db and an SSIM of 0.986, close to the leading methods.

Moreover, our method is computationally efficient with a FLOPS of 26.80G, which is significantly lower than most competing methods. This highlights our method's ability to achieve high performance without incurring substantial computational costs. These results indicate that our method is not only effective for real-world dehazing but also performs exceptionally well on synthetic datasets.

\section{B. Advantages for downstream tasks}
\label{appendix_segmentation}

Table 1 reports the performance of YOLOX detection results after dehazing by various advanced methods. Our method performs well on 4 out of 5 detection categories on the RTTS dataset, achieving the best mAP, and achieves significant performance in the key category of \textbf{pedestrian and car} detection.

\begin{table}[h]
	\linespread{1.0}
	\centering
	\setlength\tabcolsep{1.8pt} 
	\renewcommand\arraystretch{1.2}
	\scalebox{1}{	
\begin{tabular}{l|ccccccc}
\toprule
    Class(AP)  & Hazy  & PSD & C2PNet & RIDCP & DCMPNet  & Ours\\
 \midrule
 
 Bicycle & 0.618  & 0.629 & 0.624 & \textbf{0.674} & 0.655 & 0.645  \\
 
 Bus & 0.504 & 0.470 & 0.493 & 0.506  & 0.470 &\textbf{0.517}  \\
 
 Car & 0.681 & 0.734 & 0.772 & 0.782  & 0.769 &\textbf{0.797} \\
 
 Motor & 0.649  & 0.627 & 0.643 & 0.651  & 0.648 &\textbf{0.678} \\
 
 Person & 0.762  & 0.796 & 0.816 & 0.837  & 0.816 &\textbf{0.851} \\
 
\midrule
    mAP  & 0.642  & 0.651 & 0.669 & 0.690  & 0.671 &\textbf{0.697} \\
\bottomrule
 
	\end{tabular}}
    \vskip -0.05in
	\caption{Object detection results on RTTS}
	\label{table3:detection_rtts}
\end{table}

To highlight the benefits of reducing haze for subsequent tasks in real-world images, we use the Segment Anything tool for image segmentation to evaluate the advantages of different dehazing models. 

\begin{figure*}[!t]
    \centering 
    \includegraphics[width=1\textwidth]{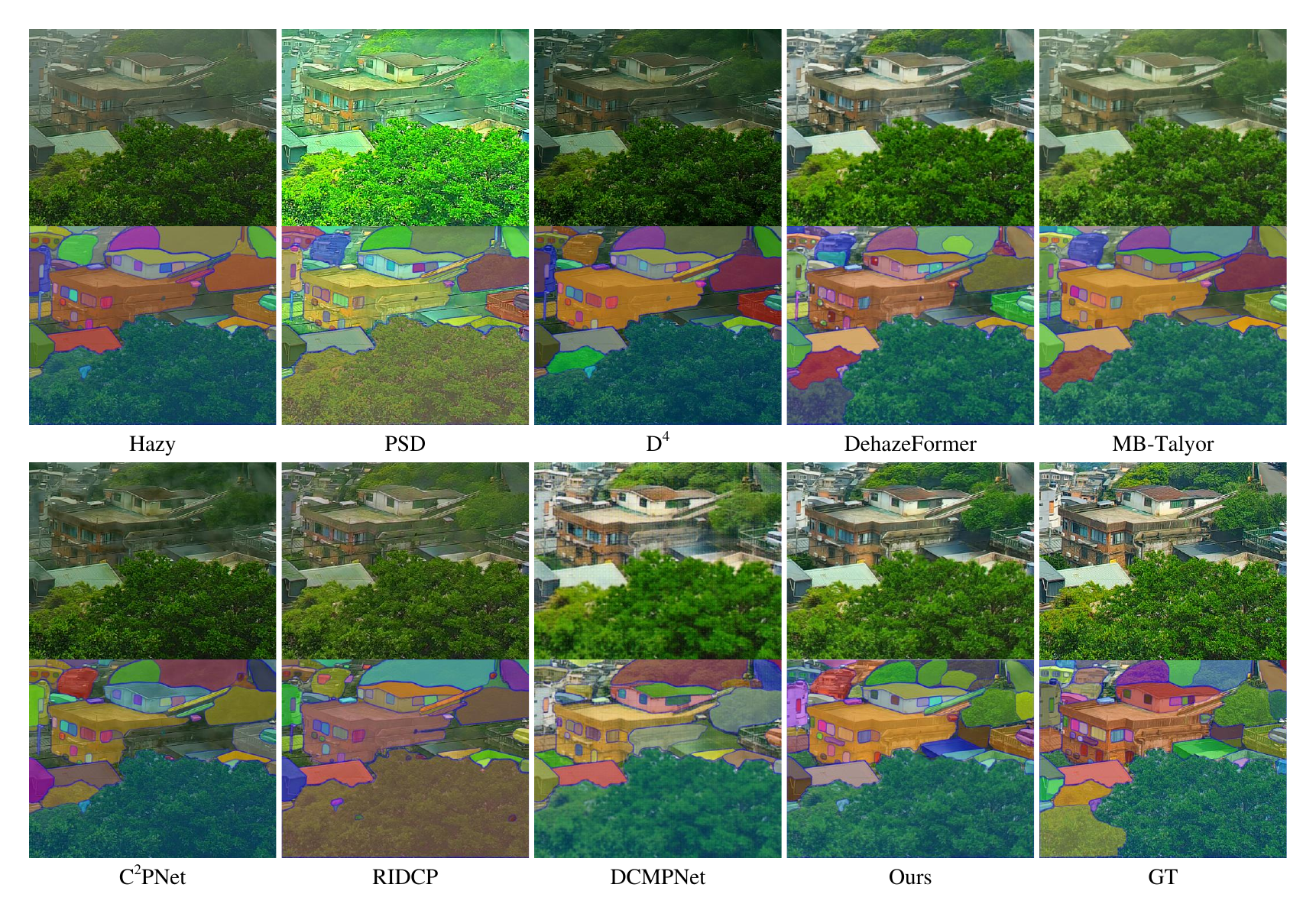}
    \vskip -0.1in
    \caption{Visual results of semantic segmentation on our RW$^2$AH dataset. Our dehazing results can help generate correct categories and continuous segmentation effects.}
    \label{figure:appendix_segmentation}
\end{figure*}

As shown in Fig. \ref{figure:appendix_segmentation}, our dehazing results demonstrate that compared to other state-of-the-art techniques, our model can more effectively segment areas with unclear boundaries. More importantly, our segmentation results are more logical and consistent with ground truth, grouping the same class into a single color. This improved performance is attributed to our SGDN's ability to restore finer texture details and scene brightness.

\textbf{Limitation:}
Here, we discuss the limitations of our SGDN model. One major challenge is handling dense, non-uniform fog patches. These patches are typically small in extent but have high density, causing sharp variations in visibility. Such scenarios make it difficult for CNN networks to extract meaningful features, primarily because the network's input lacks useful information beyond the presence of thick haze. Consequently, our model might occasionally introduce artifacts in the dehazing results.

Moreover, the variability in fog density and distribution can lead to inconsistent dehazing performance across different scenes. While our framework is robust in many real-world conditions, extreme cases of dense fog require further refinement. Future work should focus on enhancing the model's adaptability to varying fog conditions and reducing the incidence of artifacts.

\section{C. More Visual Results on Real-world Haze Dataset}
\label{appendix_more_visual}
To further validate the effectiveness of our method in real-world dehazing, we provide additional visual results on our RW$^2$AH dataset and RTTS datasets. 

\begin{figure*}[!t]
    \centering 
    \includegraphics[width=1\textwidth]{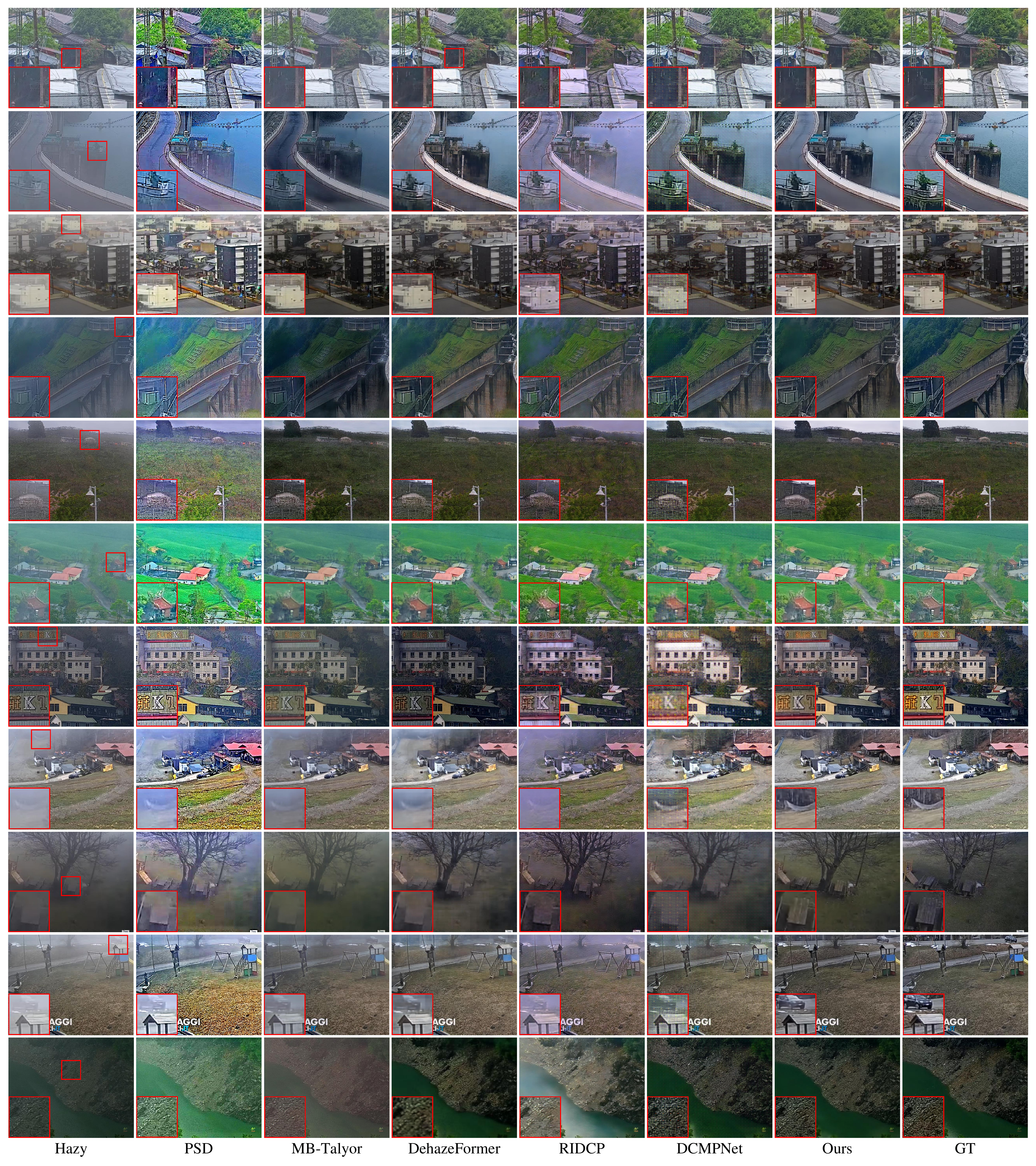}
    \vskip -0.1in
    \caption{Visual comparison results on the our real-world well-aligned haze dataset. Our results are closer to the reference images in texture and color. Zoom in for a better view.}
    \label{figure:appendix_rdhaze}
\end{figure*}

Fig. \ref{figure:appendix_rdhaze} showcases more visual comparison results on our RW$^2$AH dataset. It can be clearly demonstrating that our method effectively removes more haze while preserving natural colors. Our results in restored images that are much closer to the ground truth images.

\begin{figure*}[!t]
    \centering 
    \includegraphics[width=1\textwidth]{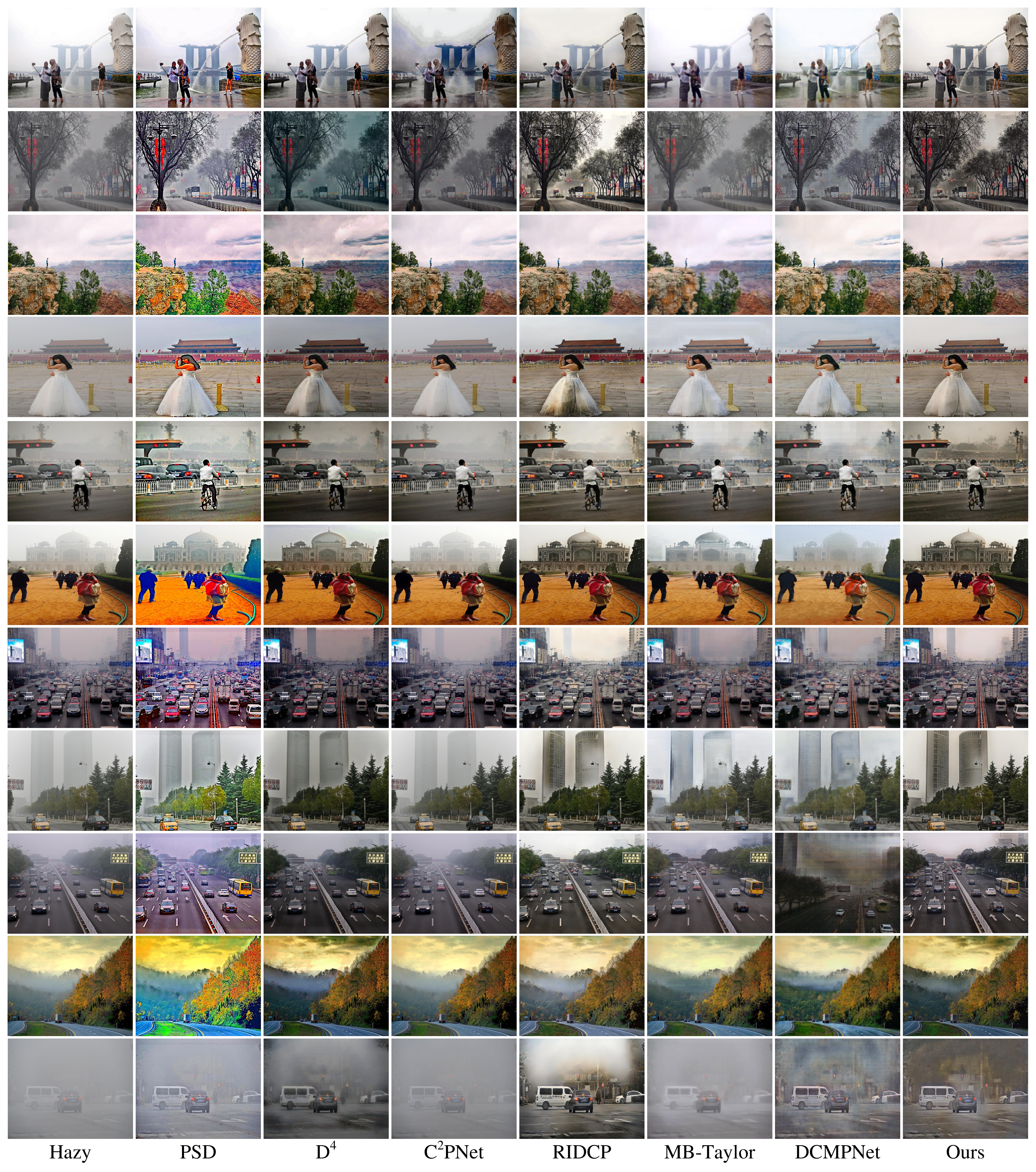}
    \vskip -0.1in
    \caption{Visual comparison results on the RTTS dataset. Our SGDN can remove more haze and maintain bright contrast. Zoom in for a better view.}
    \label{figure:appendix_rtts}
\end{figure*}

To emphasize the robustness of our SGDN, we manually selected challenging hazy images from the RTTS dataset. These images depict complex scenes with people and vehicles under dense fog, as well as distant haze conditions. We used these images to evaluate the performance of our real-world dehazing model. Fig. \ref{figure:appendix_rtts} presents a comparison of the dehazing results of various methods. The results clearly illustrate that our method outperforms the others in effectively removing haze while preserving the natural appearance of the scene. Our SGDN not only eliminates haze more comprehensively but also maintains color fidelity and image details better than competing methods. This demonstrates our method's superior ability to handle both dense and distant haze, making it more reliable and versatile for real-world applications.

\end{document}